\documentclass[10pt,twocolumn,letterpaper]{article}

\usepackage{cvpr}
\usepackage{times}
\usepackage{epsfig}
\usepackage{graphicx}
\usepackage{amsmath}
\usepackage{amssymb}

\usepackage{algorithm}
\usepackage{algorithmic}
\usepackage{multirow}

\usepackage{colortbl}
\usepackage{xcolor}

\usepackage{subfigure}

\newcommand{\tabincell}[2]{\begin{tabular}{@{}#1@{}}#2\end{tabular}}

\usepackage[pagebackref=true,breaklinks=true,letterpaper=true,colorlinks,bookmarks=false]{hyperref}

\cvprfinalcopy 

\def\cvprPaperID{1433} 

\def\w{\mathbf{w}}

\ifcvprfinal\fi
\begin{document}

\title{Adaptive Objectness for Object Tracking}


\author{\vspace{1mm}Pengpeng Liang$^1$, Chunyuan Liao$^2$, Xue Mei$^3$, and Haibin Ling$^{1}$\thanks{Correspondence author. This work is supported in part by the NSF Grant IIS-1218156 and the NSF CAREER Award IIS-1350521.}\\
\small $^1$Department of Computer and Information Sciences, Temple University, Philadelphia, PA USA\\
\normalsize $^2$HiScene Information Technologies, Shanghai, China\\
\normalsize $^3$Toyota Research Institute, North America, Ann Arbor, USA\\
{\tt\small \{pliang,hbling\}@temple.edu, liaocy@hiscene.com, xue.mei@tema.toyota.com}
}

\maketitle

\begin{abstract}
 Object tracking is a long standing problem in vision. While great efforts have been spent to improve tracking performance, a simple yet reliable prior knowledge is left unexploited: the target object in tracking must be an \emph{object} other than non-object. The recently proposed and popularized \emph{objectness} measure provides a natural way to model such prior in visual tracking. Thus motivated, in this paper we propose to adapt objectness for visual object tracking. Instead of directly applying an existing objectness measure that is generic and handles various objects and environments, we adapt it to be compatible to the specific tracking sequence and object. More specifically, we use the newly proposed BING~\cite{cheng2014bing} objectness as the base, and then train an object-adaptive objectness for each tracking task. The training is implemented by using an adaptive support vector machine that integrates information from the specific tracking target into the BING measure. We emphasize that the benefit of the proposed adaptive objectness, named ADOBING, is generic. To show this, we combine ADOBING with seven top performed trackers in recent evaluations. We run the ADOBING-enhanced trackers with their base trackers on two popular benchmarks, the CVPR2013 benchmark (50 sequences) and the Princeton Tracking Benchmark (100 sequences). On both benchmarks, our methods not only consistently improve the base trackers, but also achieve the best known performances. Noting that the way we integrate objectness in visual tracking is generic and straightforward, we expect even more improvement by using tracker-specific objectness.
\end{abstract}

\section{Introduction}
\label{sec:introduction}
Visual object tracking is a fundamental computer vision task with a wide line of applications for human-computer interaction, surveillance, vehicle navigation, etc. Various factors, including illumination changes, partial occlusions, pose variations and background clutter, challenge tracking algorithms in practice. To handle these factors, a great amount of efforts have been devoted to develop robust observation model by utilizing the local structure of the target and/or visual cues such as shape and appearance.


\begin{figure}[t]
\centering
    \includegraphics[width=0.495\linewidth, height=0.33\linewidth]{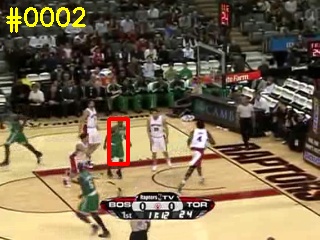}
    \hfill
    \includegraphics[width=0.495\linewidth, height=0.33\linewidth]{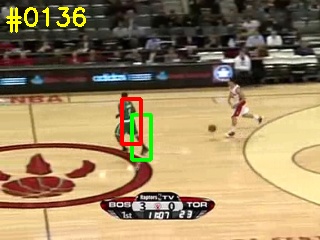}
    \\
\vspace{1mm}
    \includegraphics[width=0.495\linewidth, height=0.33\linewidth]{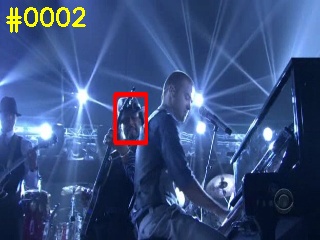}
    \hfill
    \includegraphics[width=0.495\linewidth, height=0.33\linewidth]{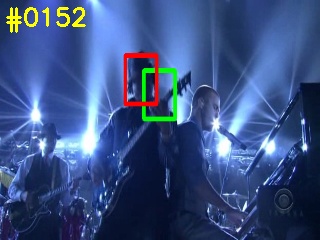}
\caption{Improving tracking by integrating objectness. The results on the right frames show that when the base tracker (in green, Struck~\cite{hare2011struck}) starts drifting, the proposed objectness-aware solution (in red, Struck+ADOBING) successfully avoids such pitfalls. }
\label{fig:highlight-success}
\end{figure}

Despite a large amount of previous efforts, little attention has been given to a simple yet reliable prior that a visual target under tracking should first be an \emph{object} rather than not. An obvious advantage of integrating such information is to inhibit drifting, as observed in our experiments (e.g., Fig.~\ref{fig:highlight-success}). This intuition naturally directs our attention to the recently proposed and popularized
\emph{objectness} measure~\cite{alexe2012measuring} that estimates the likelihood that a given image window contains a whole object.

To apply the objectness for visual object tracking, however, there are two issues need to be addressed. The first is speed, for which we luckily have the newly developed fast objectness algorithm named BING~\cite{cheng2014bing}. The second one is adaptivity, since objectness is originally designed to handle generic objects under various environment; while for tracking, we typically focus on a specific object in a relative stable environment.



Guided by the above idea, in this paper we propose to integrate objectness for object tracking. First, we derive a novel \emph{adaptive objectness} based on BING, named ADOBING, that adapts BING to a specific tracking task. In particular, given an image sequence and the initial object-of-interest, ADOBING is learnt by an adaptive SVM that adjusts BING according to the tracking object and background in the initial frame. This way, the generic objectness is effectively balanced with a specific tracking task at hand, meanwhile the extreme computational efficiency of BING is inherited.

We then integrate ADOBING to existing visual trackers to show the general advantage of using objectness for tracking. Towards the goal, instead of designing a specific mechanism to improve a specific tracker, we employ a straightforward strategy, {\it i.e.}, linear combination of the original tracking confidence and ADOBING, to improve seven trackers (called \emph{base trackers}) that have achieved top performances in recent tracking benchmarks~\cite{wu2013online,pang2013finding,kristanvot2013}.
We test the objectness enhanced trackers on two recently proposed tracking benchmarks: the CVPR2013 tracking benchmark~\cite{wu2013online} and the Princeton Tracking Benchmark~\cite{song2013tracking}. The results show not only the consistent improvement over the base trackers by using objectiveness, but also the advantage of adapting the original objectness to a tracking specific one. In addition, with the help of objectness, we have achieved the best results ever reported on these benchmarks.

In summary, our contributions are three-fold: (1) integrating objectness for visual tracking, (2) developing a tracking-adaptive objectness, and (3) thorough experimental validation with state-of-the-art performance.

In the rest of the paper, we first summarize related work in Sec.~\ref{sec:related}. Then, we introduce the proposed adaptive objectness and its integration in tracking in Sec.~\ref{sec:approach}. The experimental validation is described in Sec.~\ref{sec:exp}, followed by conclusion in Sec.~\ref{sec:conclusion}.

\section{Related work}
\label{sec:related}

\vspace{-2.001mm}\paragraph{Visual Object Tracking.} Visual tracking has been studied for several decades and it is beyond this paper to give a comprehensive review. Surveys of tracking algorithms can be found in \cite{yilmaz2006object}, or in tracking evaluation papers~\cite{wu2013online,pang2013finding,smeulders2013visual,vot2014} for more recent progresses. In the following we review some most related works.


While image-based objectness is new for visual tracking, a related concept, namely visual saliency, has been recently connected to visual tracking in~\cite{mahadevan2013biologically,mahadevan2012connections,su2014abrupt}. In particular, a biologically inspired object tracking algorithm was proposed in  \cite{mahadevan2013biologically}, which selects the most informative features by utilizing the connection between discriminant saliency and the Bayes error for target/background classification. The relationship between tracking reliability and the degree target saliency was studied in~\cite{mahadevan2012connections} by human behavior experiments. In~\cite{su2014abrupt}, to deal with abrupt motion, the target is relocated by searching the salient region obtained from an adaptive saliency map when the target gets lost.  It is also worth noting that the work in~\cite{stalder2013dynamic} uses a motion saliency mechanism which considers the specific motion of the target was developed to re-discover the tracked object.
Despite related to our work, none of the above studies takes the prior into account that a tracking target needs to be an object. As far as we know, our work is the first to explicitly model such prior for visual object tracking.


\vspace{-2.001mm}\paragraph{Objectness.} The concept of objectness is first proposed by Alexe et al.~\cite{alexe2012measuring} to reflects the likelihood that an image window contains an entire object. The objectness estimator is trained using various image cues, such as multi-scale saliency, color contrast, edge density and superpixel straddling, to model regions that stand out from the surroundings and have a closed boundary.
The objectness thus learned is generic and can be applied to many vision tasks for improving accuracy or speed, or both. As a result, it has been recently applied to a series of tasks such as object detection~\cite{ChangLCL11iccv}, image retargeting~\cite{Sun&Ling13ijcv}, action localization \cite{jain2014action}, salient region segmentation~\cite{JiangLYP13iccv}, scene classification \cite{juneja2013blocks}, image retrieval \cite{tao2014locality}, etc.

Ideas similar to objectness has also been developed for vision tasks. For example, selective search based on hierarchical grouping with diversified criteria was proposed in \cite{uijlings2013selective} to generate high quality object proposals.
In~\cite{zhang2011proposal}, cascaded ranking SVMs were used for object proposal generation for object detection. In~\cite{endres2013category}, regions from segmentation can be further ranked with structure learning to produce object proposals. 

Being effective for many vision tasks, the computation cost of the original objectness forbids it from tasks that request high speed responses such as tracking. Addressing this problem, Cheng et al.~\cite{cheng2014bing} broke through the computational bottleneck by proposing a very fast objectness measure named BING (a short review is given in Sec.~\ref{sub:bing}). The extreme high efficiency (300 fps on a laptop) opens a way for using BING in many real-time applications and we adopt BING as the basis in our study.
The objectness proposed in this paper builds on top of BING by automatically adapting it for specific visual tracking tasks. This way, our new objectness measure enjoys both fast objectness inference and tracking-oriented accuracy improvement.


\begin{figure*}[!t]
\includegraphics[width=1.0\linewidth, height=0.4\linewidth]{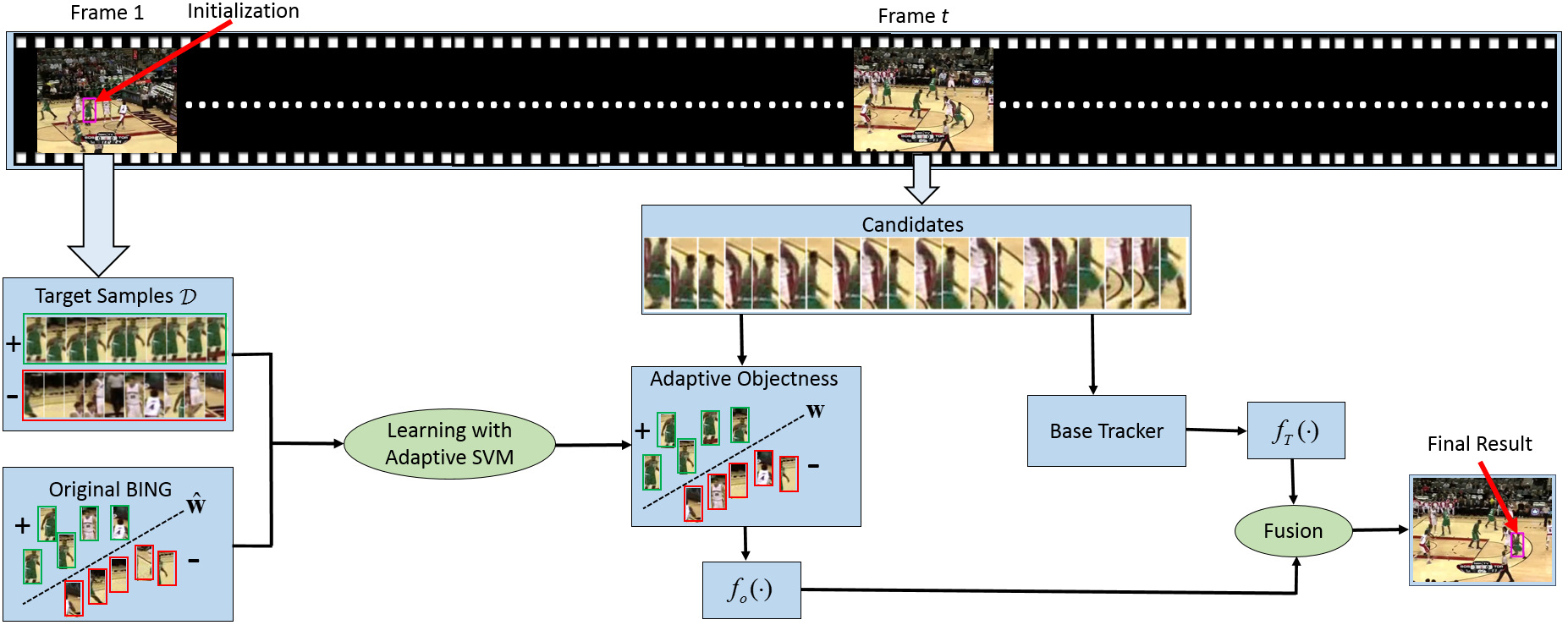}
\caption{Framework of integrating adaptive-objectness for object tracking.}
\label{fig:pipeline}
\end{figure*}

\vspace{-2.001mm}\paragraph{Model transfer/adaptation for tracking.} Transfer learning deals with tasks where the target task can benefit from adapting knowledge or models learned from training samples that have a distribution correlated with yet different from the distribution of the testing samples~\cite{pan2010survey}. In our study, we apply transfer learning to SVM-based classifier which is used in our base objectness model ({\it i.e.}, BING). In this aspect, our work is closely related to and inspired by the work in~\cite{yang2007adapting,tommasi2013learning}. In particular, to increase the amount of transfer without penalizing the margin, projective model transfer SVM was proposed in \cite{aytar2011tabula}, and the regularization term was further extended for deformable source template. The adapt-SVM~\cite{yang2007adapting} modifies the regularization term for the model parameters of SVM to transfer knowledge from a single already learned source model. In~\cite{tommasi2013learning}, in order to transfer knowledge from multiple source models and alleviate negative transfer, an adaptive least-square support vector machine was proposed which could weight the source models differently.



Transfer learning has been applied to visual tracking in several previous studies~\cite{gao2014transfer,li2011treat,wang2012transferring,wang2013deep}. In \cite{li2011treat}, a semi-supervised online boosting algorithm based on ''Covariate shift`` is proposed for tracking. The approaches in \cite{wang2012transferring,wang2013deep} use prior knowledge learned offline from real-world natural images to represent image patches. In \cite{gao2014transfer}, prior knowledge learned by online regression on auxiliary examples is transferred to assist the final decision.   Our work is different in two main aspects: (1) we apply transfer learning to objectness instead of directly on tracking; and (2) we use $\ell_1$ regularized adaptive SVM which is different than in previous studies.

\section{Approach}
\label{sec:approach}

\subsection{Overview}
\label{sub:overview}

Our proposed framework to enhance visual tracking by adaptive objectness is summarized in
Figure~\ref{fig:pipeline}. In the following we give a brief description, and postpone more details in other subsections.

Given an input image sequence and tracking initialization (e.g., bounding box for object-of-interest), our framework starts by first learning the adaptive objectness, i.e., ADOBING, using an adaptive SVM. The learning takes two components as input: the tracking-specific training samples $\mathcal{D}$ extracted from the tracking initialization, and the base BING objectness algorithm represented by its parameter vector $\mathbf{\widehat{w}}$ from~\cite{cheng2014bing}.

Let $T$ be the base tracker to be enhanced, $f_T(\cdot)$ be the tracking confidence of $T$ for tracking candidates and $f_O(\cdot)$ be the learned ADOBING objectness.  Then, during tracking, for each tracking candidate $\mathbf{c}$, we fuse its base tracking inference $f_T(\mathbf{c})$ and its adaptive objectness $f_O(\mathbf{c})$ in a weighted linear combination. The tracking result is then selected according the fused confidence.


\subsection{Object-adaptive Objectness}
\label{sec:oao}

\subsubsection{Review of BING}
\label{sub:bing}
In \cite{cheng2014bing}, a 64D \emph{binarized normed gradients} (BING) feature was proposed for efficient objectness estimation. Motivated by the fact that objects are stand-alone things with well-defined closed boundaries and centers~\cite{alexe2012measuring,forsyth1996finding}, BING first resizes image windows to a small fixed size ($8\times8$ is chosen for the computational reason), then uses the corresponding normed gradients to discriminate objects and non-object stuff in an image.

In the training stage, it trains a linear model $\mathbf{\widehat{w}}\in \mathbb{R}^{64}$ with linear SVM. In the testing stage, the model $\mathbf{\widehat{w}}$ is approximated with $N_w$ binary vectors $\mathbf{a}^{+}_{j}$ and their complements $\overline{\mathbf{a}_{j}^{+}}$ weighted by $\beta_{j}$.  The 64D normed gradients (each element is saved as a {\sc BYTE} value) is approximated by $N_g$ binarized normed gradients (BING) feature as $\mathbf{g}=\sum_{k=1}^{N_g}2^{8-k}\mathbf{b}_k$, where $\mathbf{b}_k\in\{0,1\}^{64}$ and is the binary approximation of $\mathbf{g}$ at the $k_{\text{th}}$ bit. Then, the confidence score of an image window can be efficiently estimated using:
\begin{equation*}
    s\approx\sum_{j=1}^{N_w}\beta_{j}\sum_{k=1}^{N_{g}}C_{j,k}
\end{equation*}
where $C_{j,k}=2^{8-k}(2\langle\mathbf{a}^{+}_{j},\mathbf{b}_k\rangle-|\mathbf{b}_k|)$. Since the dimension of $\mathbf{a}^{+}_{j}$ and $\mathbf{b}_k$ is 64, they can be stored with {\sc int64}, and $C_{j,k}$ can be tested using fast {\sc bitwise} and {\sc popcnt sse} operators.

\subsubsection{Learning Adaptive Objectness}
\label{sec:tsobj}

\begin{figure}[!t]
\centering
\subfigure[Tracking input]{
\label{fig:show-couple}
\includegraphics[width=0.32\linewidth, height=0.23\linewidth]{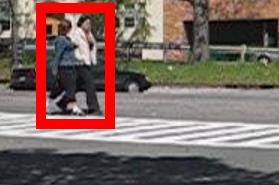}
}
\hspace{-8pt}
\subfigure[BING]{
\label{fig:show-couple-BING}
\includegraphics[width=0.32\linewidth, height=0.23\linewidth]{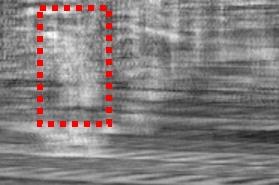}
}
\hspace{-8pt}
\subfigure[ADOBING]{
\label{fig:show-couple-OAO}
\includegraphics[width=0.32\linewidth, height=0.23\linewidth]{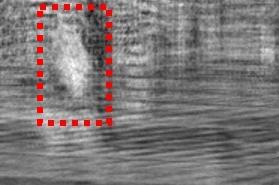}
}
\caption{Objectness (BING) and adaptive objectness (ADOBING) for a specific tracking task. (a) The first frame with the initialization of the tracking target (red bounding box). (b) the objectness map of BING. (c) The objectness map of ADOBING.} 
\label{fig:OAO-BING-compare}
\end{figure}

As briefed in Sec.~\ref{sub:overview}, we can formulate the learning of adaptive objectness as following: Given the training data $\mathcal{D}=\{\mathbf{x}_i,y_i\}_{i=1}^{N}$ and a previously learned linear model $\mathbf{\widehat{w}}\in \mathbb{R}^{64}$ ({{\it i.e.}, BING), where $\mathbf{x}_i\in \mathbb{R}^{64}$ is a normed gradients vector of an image patch and $y_i=\pm1$ is its binary label, our task is to learn a linear model ${\w}$ adapted from $\widehat{{\w}}$.

\vspace{-2.001mm}\paragraph{Objective function.} To train the linear model ${\w}$, we use the adaptive SVM framework~\cite{yang2007adapting} so that the discrepancy between ${\w}$ and $\mathbf{\widehat{w}}$ can be constrained while minimizing the classification error over $\mathcal{D}$.  Specifically, the regularizer $\|{\w}\|_1$ in standard $\ell_1$-regularized linear SVM \cite{yuan2010comparison} is replaced by $\|{\w}-\mathbf{\widehat{w}}\|_1$, and resulting the following objective function:
\begin{equation}
\label{eq:asvm}
    \mathop{\min_{{\w}}}\|{\w}-\mathbf{\widehat{w}}\|_1+ C\sum_{i=1}^{N}(\max(0,1-y_i{\w}^{T}\mathbf{x}_i))^2 ~,
\end{equation}
where $C$ is the regularization weight.

\vspace{-2.001mm}\paragraph{Solving (\ref{eq:asvm}) with Coordinate Descent.}
We employ the coordinate descent algorithm with one-dimensional Newton direction to solve the optimization in (\ref{eq:asvm}). The idea is to iteratively improve $\w$, and, in each iteration, improving $\w$ sequentially along each dimension. In the following, we use $\w^{(k)}$ for $\w$ at the beginning of the $k$-th iteration, and $\w^{(k,j)}$ for $\w^{(k)}$ after updating along its $j$-th dimension ($1\le j\le 64$).

Following~\cite{fan2008liblinear}, for the $i$-th training sample $(\mathbf{x}_i,y_i)$, we define $b_i(\w)\equiv 1-y_i{\w}^\top\mathbf{x}_i$ and $I({\w})\equiv\{i|b_i(\w)>0\}$. Then, a coordinate descent for updating the $j$-th component of $\w^{(k)}$ is achieved by solving the following one-dimensional sub-problem:
\begin{equation}\label{eq:subprob}
\begin{split}
   \mathop{\min_{z}}  g_j(z) = &|{\w}^{(k,j)}_j-\mathbf{\widehat{w}}_j+z|-|{\w}^{(k,j)}_j-\mathbf{\widehat{w}}_j|\\
   & + L_j(z;{\w}^{(k,j)}) - L_j(0;{\w}^{(k,j)})
\end{split}
\end{equation}
where subscript $j$ indicates the $j$-th element of the associated vector, and  $L_j(z;\mathbf{u})=C \sum_{i\in I(\mathbf{u}+z\mathbf{e}_j)} b_i(\mathbf{u}+z\mathbf{e}_j)^2$, and $\mathbf{e}_j\in\mathbb{R}^{64}$ is the vector with the $j$-th element be $1$ and all others be $0$. The coordinate descent framework is summarized in Algorithm \ref{alg:cdm}.

\begin{algorithm}[!t]
\caption{Coordinate Descent~\cite{yuan2010comparison}}
\label{alg:cdm}
\begin{algorithmic}[1]
\REQUIRE ${\w}^{(1)}\in\mathbb{R}^n$
\FOR{k=1,2,..., iterate until convergence}
    \STATE ${\w}^{(k,1)}={\w}^{(k)}$
    \FOR{$j=1,2,\cdots,n$}
        \STATE Find $z$ by solving the sub-problem (\ref{eq:subprob}) exactly or approximately.
        \STATE ${\w}^{(k,j+1)}={\w}^{(k,j)}+z\mathbf{e}_j$.
    \ENDFOR
    \STATE ${\w}^{(k+1)} = {\w}^{(k,n+1)}$.
\ENDFOR
\STATE \textbf{return} $\w^{(k)}$
\end{algorithmic}
\end{algorithm}

Note that $g_j(z)$ is not differentiable, to solve the (\ref{eq:subprob}), we calculate the Newton direction by considering only the second-order approximation of the loss term $L_j(z;{\w}^{(k,j)})$ and solve
\begin{equation}\label{eq:secondorder}
\begin{split}
   \mathop{\min_{z}}~ & |{\w}^{(k,j)}_j-\mathbf{\widehat{w}}_j+z|-|{\w}^{(k,j)}_j-\mathbf{\widehat{w}}_j|\\
   & + L_j^{\prime}(0;{\w}^{(k,j)})z +\frac{1}{2}L_j^{\prime\prime}(0;{\w}^{(k,j)})z^2 ,
\end{split}
\end{equation}
where
\begin{equation}\label{eq:firstL}
    L_j^{\prime}(0;{\w}^{(k,j)})=-2C\sum_{i\in I({\w}^{(k,j)})}y_i\mathbf{x}_{ij}b_i({\w}^{(k,j)}) ,
\end{equation}
and
\begin{equation}\label{eq:secondL}
  L_j^{\prime\prime}(0,{\w}^{(k,j)})=2C\sum_{i\in I({\w}^{(k,j)})}\mathbf{x}_{ij}^2 ~.
\end{equation}
Note that $L_j^{\prime\prime}(0,{\w}^{(k,j)})$ in the above formula is a generalized second derivative \cite{chang2008coordinate}, since $L_j(z;{\w}^{(k,j)})$ is not twice differentiable.

With some derivation similar in \cite{yuan2010comparison}, it can be shown that (\ref{eq:secondorder}) has the following closed-form solution:
\begin{equation}\label{eq:newton-direction}
    d = \left\{
    \begin{array}{ll}
    \vspace{1.5mm} -\frac{L_j^{\prime}(0;{\w}^{(k,j)})+1}{L_j^{\prime\prime}(0;{\w}^{(k,j)})} & \text{ if } s_j({\w}^{(k,j)})\leq -1,\\
    \vspace{1.5mm} -\frac{L_j^{\prime}(0;{\w}^{(k,j)})-1}{L_j^{\prime\prime}(0;{\w}^{(k,j)})} & \text{ if } s_j({\w}^{(k,j)})\geq 1\\
    \mathbf{\widehat{w}}_j - {\w}^{(k,j)}_j & \text{ otherwise}.
    \end{array}
    \right.
\end{equation}
where $s_j(\mathbf{v})=L_j^{\prime}(0;\mathbf{v})-L_j^{\prime\prime}(0;\mathbf{v})(\mathbf{v}_j-\mathbf{\widehat{w}}_j)$.

We then conduct a line search procedure to check if $\beta^{t}d$ satisfy the following sufficient decrease condition:
\begin{equation}\label{eq:line-search-condition}
\begin{split}
    g_j(\beta^t d)-&g_j(0) \leq \sigma\beta^t\Big( L_j^{\prime}(0;{\w}^{(k,j)}) d \\
        &   + |\w^{(k,j)}_j-\mathbf{\widehat{w}}_j+d|   - |\w^{(k,j)}_j-\mathbf{\widehat{w}}_j| \Big)
\end{split}
\end{equation}
where $\beta\in(0,1)$, $t=0,1,2,\cdots$, and $\sigma\in(0,1)$. The first $\beta^td$ that satisfies the condition (\ref{eq:line-search-condition}) is chosen as the solution for the sub-problem (\ref{eq:subprob}).


\subsection{Encoding Adaptive Objectness for Tracking}
\label{sec:encoding}

As illustrated in Fig.~\ref{fig:pipeline}, in addition to the learning algorithm, three components are needed for encoding objectness for tracking, including preparing training samples, selecting a base tracker, and fusing the base tracker with the proposed adaptive objectness.

\vspace{-2.001mm}\paragraph{Generating training samples $\mathcal{D}$.}
For each sequence, we use the first frame to generate training samples $\mathcal{D}$ with a sliding window fashion over the entire image. One image patch is labeled as positive if its overlap with the ground truth is greater than some predefined threshold; otherwise, it is labeled as negative.
Figure~\ref{fig:OAO-BING-compare} presents the confidence map of the original BING and the learned adaptive objectness (ADOBING) for a specific tracking task.

In this study we limit the samples to the first frames mainly for two reasons. First, theoretically, only the initialization is guaranteed to be the true target and tracking results from the second frame can be polluted. Second, though we aim to adjust the original objectness for the specific tracking sequence, we want to avoid overfitting the objectness. In other words, using limited number of samples balances the generic property and the tracking specificity of the proposed adaptive objectness. That been said, in practice, one may collect more samples from several initial frames for improvement.

\vspace{-2.001mm}\paragraph{Selecting base trackers.}
It is impractical to use all existing tracking algorithms to validate the efficacy of integrating objectness, instead we select top ranked trackers in recent tracking evaluations~\cite{wu2013online,pang2013finding,kristanvot2013}. More specifically, we first create an initial set of trackers that ranked within top 10 in any of these evaluations. Then, from these trackers, we chose those with open source implemented in C/C++, since BING and our ADOBING are both implemented in C++. Such selection provides us six base trackers, namely BSBT~\cite{stalder2009beyond}, Frag~\cite{adam2006robust}, MIL~\cite{babenko2011robust}, OAB~\cite{grabner2006real}, SemiT~\cite{grabner2008semi}, and Struck~\cite{hare2011struck}.

Furthermore, there are several recently proposed trackers with higher reported performance on the above mentioned benchmarks, such as~\cite{gao2014transfer,henriques2014high}. From such trackers, we select TGPR~\cite{gao2014transfer} since it has C/C++ implementation available.

In summary, we have seven state-of-the-art trackers selected as base trackers. As will be clear in Sec.~\ref{sec:exp}, though they have already achieved top performances in previous benchmark evaluations, their performances can still be boosted by integrating the proposed adaptive objectness and produce best results known so far.

\vspace{-2.001mm}\paragraph{Encoding objectness.}
Given a base tracker denoted by $T$, one may improve it by integrating objectness in a tracker-specific way so as to maximize the benefit from objectness. However, in this paper, we are more interested in showing that the benefit provided by objectness is generic. Therefore, we follow a straightforward strategy to directly combine the tracking confidence from $T$ with the objectness measure. This strategy is very general and applicable to all selected base trackers as well as most other modern trackers.

Roughly speaking, for a base tracker $T$, when a new frame arrives, to identify the tracking target, a set of candidate $\mathcal{C}=\{\mathbf{c}_i\}$ is first constructed; then a tracking confidence $f_T(\cdot)$ applies to each candidate; finally the candidate with maximum confidence value is selected as the tracking result.  To integrate objectness (either BING or ADOBING), we simply replace $f_T$ by an objectness-enhanced confidence $f_{OT}$, and $f_{OT}$ is a weighted sum of $f_T$ and $f_O$, where $f_O(\cdot)$ is the objectness measure of a candidate. In particular, for a candidate $\mathbf{c}_i\in\mathcal{C}$, we have
\begin{equation}\label{eq:confidence}
    f_{OT}(\mathbf{c}_i) = f_T(\mathbf{c}_i) + \lambda f_O(\mathbf{c}_i) ~,
\end{equation}
where $\lambda$ is a constant weight.

The above strategy has been applied to all seven base trackers. For each base tracker, we normalize the original confidence (probability, cost, etc.) for the strategy. It is emphasizing again that, despite its simplicity, the strategy boosts consistently all base trackers in our experiment when using the proposed adaptive objectness (ADOBING).


\begin{figure*}[t]
\centering
    \includegraphics[width=0.495\linewidth,height=0.31\linewidth]{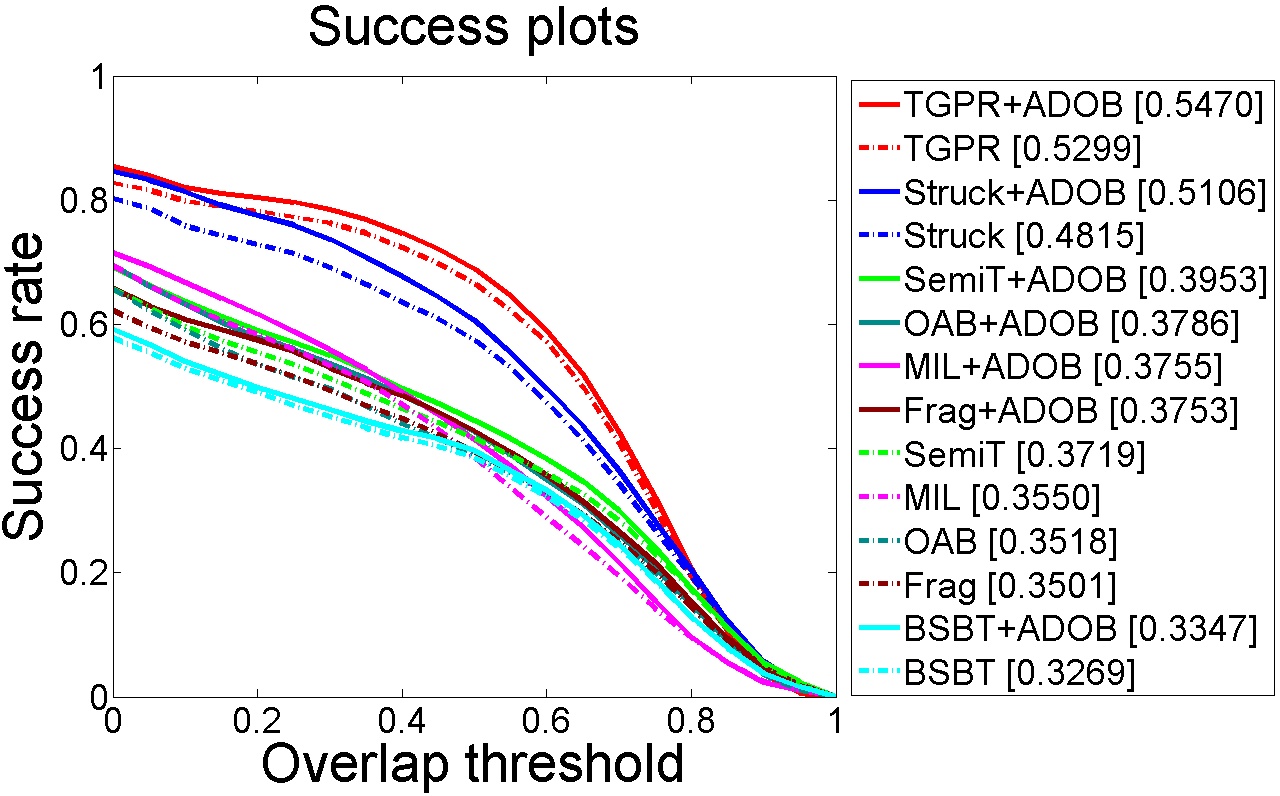}
    \hfill \includegraphics[width=0.495\linewidth,height=0.31\linewidth]{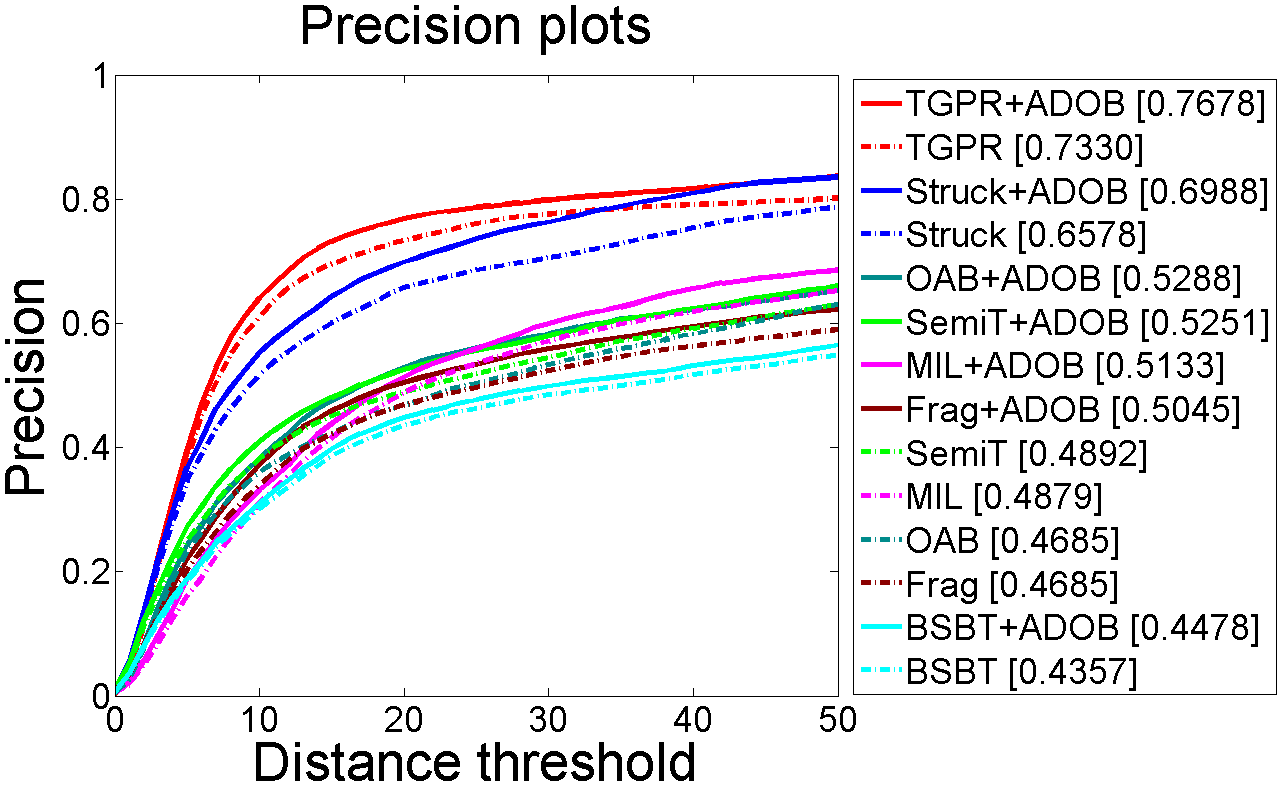}
    \caption{Success and precision plots for all ADOBING-enhanced trackers and base trackers on the CVPR2013 benchmark.}
\label{fig:trackers-compare-cvpr2013}
\end{figure*}

\begin{table*}[!t]
\small
\caption{The performance of BING- or ADOBING-enhanced trackers and their corresponding base trackers in term of AUC on the CVPR2013 benchmark. The \textcolor{red}{\bf best} and \textcolor{blue}{second best} for each tracker are indicated by \textcolor{red}{\bf red} and \textcolor{blue}{blue} respectively.}
\centering
\begin{tabular}{|c|c|c|c|c|c|c|c||c|}
\hline
&BSBT \cite{stalder2009beyond}&Frag \cite{adam2006robust}&MIL \cite{babenko2011robust}&OAB \cite{grabner2006real}&SemiT \cite{grabner2008semi}&Struck \cite{hare2011struck}&TGPR \cite{gao2014transfer}&Average\\
\hline
Base&\textcolor{blue}{0.327}&0.350&0.355&0.352&0.372&0.482&0.530&0.395\\
\hline
Base+BING&0.320&\textcolor{blue}{0.374}&\textcolor{blue}{0.358}&\textcolor{blue}{0.360}&\textcolor{blue}{0.382}&\textcolor{blue}{0.496}&\textcolor{blue}{0.545}&\textcolor{blue}{0.405}\\
\hline
Base+ADOB&\textcolor{red}{\bf 0.335}&\textcolor{red}{\bf 0.375}&\textcolor{red}{\bf 0.375}&\textcolor{red}{\bf 0.379}&\textcolor{red}{\bf 0.395}&\textcolor{red}{\bf 0.511}&\textcolor{red}{\bf 0.547}&\textcolor{red}{\bf 0.417}\\
\hline
\end{tabular}
\label{table:auc-compare-cvpr2013}
\end{table*}

\section{Experiments}
\label{sec:exp}

We evaluate the proposed objectness-enhanced tracking algorithms on two recently published benchmarks: the CVPR2013 benchmark~\cite{wu2013online} and the Princeton Tracking Benchmark~\cite{song2013tracking}.  For the parameter setting of the base trackers, we set them as the default. For adaptive SVM, we set $C=0.01$ in Eq.(\ref{eq:asvm}); for combining the confidence from a base tracker with the objectness measure, we set $\lambda=0.1$ in Eq.(\ref{eq:confidence}).  These parameter settings are throughout all the experiments.

In the experiment, in addition to test each base tracker along with its ADOBING-enhanced version, we also run a BING-enhanced version that uses the original BING objectness. In the following, for a base tracker ``T", we use ``T+ADOB" and ``T+BING" to denote the two objectness-enhanced version of ``T". 


\begin{table*}[!t]
\small
\caption{The performance (AUC) gain under different challenging factors. The attributes are ordered according to the gain.}
\centering
\begin{tabular}{|c|c|c|c|c|c|c|c|c|c|c|c|}
\hline
&OV&MB&FM&IPR&DEF&OCC&OPR&IV&BC&SV&LR\\
\hline
&&&&&&&&&&&\\[-3.5mm]
\hline
Base&0.370&0.351&0.362&0.381&0.364&0.336&0.369&0.371&0.380&0.348&0.301\\
\hline
Base+ADOB&0.405&0.376&0.387&0.403&0.386&0.358&0.390&0.390&0.397&0.361&0.312\\
\hline
&&&&&&&&&&&\\[-3.5mm]
\hline
Gain&0.035&0.025&0.025&0.022&0.022&0.022&0.021&0.019&0.017&0.013&0.011\\
\hline
\end{tabular}
\label{table:att-gain-cvpr2013}
\end{table*}

\subsection{CVPR2013 Visual Tracking Benchmark}
The CVPR2013 Visual Tracking Benchmark \cite{wu2013online} includes 50 fully annotated sequences.  To further understand the strength and weakness of tracking algorithms, these sequences are categorized according to 11 challenging factors containing illumination variation (IV), scale variation (SV), occlusion (OCC), deformation (DEF), motion blur (MB), fast motion (FM), in-plane rotation (IPR), out-of-plane rotation (OPR), out-of-view (OV), background clutter (BC), and low resolution (LR). 

We follow the protocol in~\cite{wu2013online} for evaluation. One metric is the \emph{center location error} (CLE), defined as the Euclidean distance between the center of the tracked target position and the manually labeled ground truth. The average CLE over all frames can be used to measure the performance for that sequence. However, such measurement is not meaningful when the tracker loses the target completely. \emph{Precision plot} addresses this issue by showing the percentage of frames whose CLEs are within a given threshold. As in~\cite{wu2013online}, the precision score at the threshold = 20 pixels is used to rank the trackers in our evaluation.

Another widely used metric is based on the bounding box overlap. For each frame, given the tracking output bounding box ($r_t$) and the ground truth bounding box ($r_g$), the overlap score $S=\frac{|r_t\cap r_g|}{|r_t\cup r_g|}$ is used to measure tracking success, where $|\cdot|$ denotes the area. To quantize the tracking performance of a tracker on a sequence of frames,  we calculate the percentage of frames whose overlap score is larger than a given threshold. The \emph{success plot} can then be generated by varying the threshold from $0$ to $1$. The \emph{Area Under Curve} (AUC) derived from the success plot is used to rank the trackers. Comparing with the precision obtained at the threshold 20, AUC measures the overall performance and is therefore more accurate, so we mainly use AUC in our analysis.

\begin{figure*}[!t]
\footnotesize
\centering
\includegraphics[width=0.16\linewidth,height=0.11\linewidth]{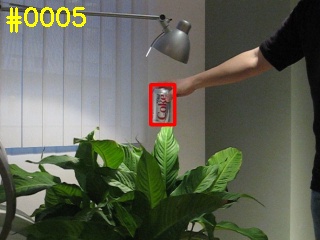}
\hspace{-2pt}
\includegraphics[width=0.16\linewidth,height=0.11\linewidth]{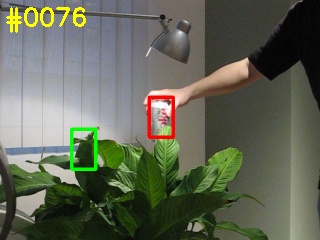}
\hspace{-2pt}
\includegraphics[width=0.16\linewidth,height=0.11\linewidth]{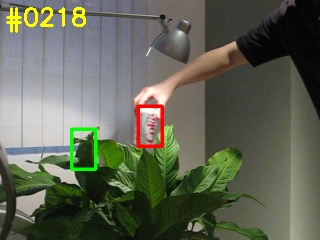}
\hspace{-0pt}
\includegraphics[width=0.16\linewidth,height=0.11\linewidth]{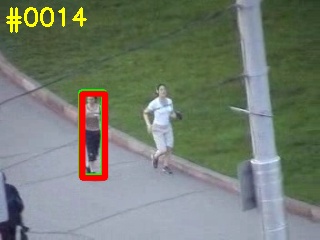}
\hspace{-2pt}
\includegraphics[width=0.16\linewidth,height=0.11\linewidth]{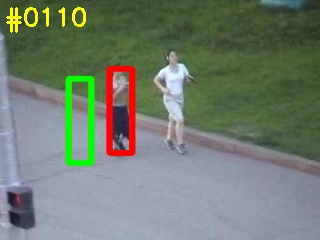}
\hspace{-2pt}
\includegraphics[width=0.16\linewidth,height=0.11\linewidth]{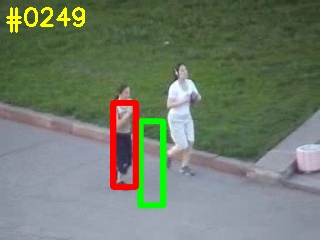}\\
(a) BSBT \hspace{210pt} (b) Frag\\
\vspace{4pt}
\includegraphics[width=0.16\linewidth,height=0.11\linewidth]{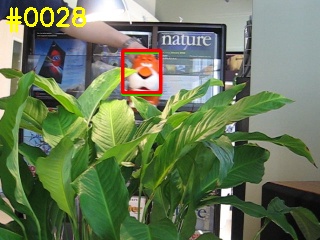}
\hspace{-2pt}
\includegraphics[width=0.16\linewidth,height=0.11\linewidth]{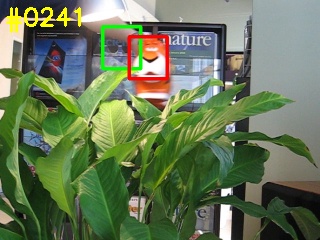}
\hspace{-2pt}
\includegraphics[width=0.16\linewidth,height=0.11\linewidth]{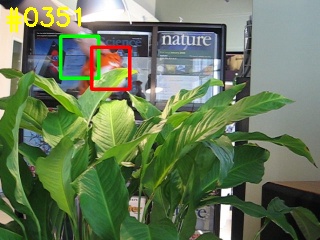}
\hspace{-0pt}
\includegraphics[width=0.16\linewidth,height=0.11\linewidth]{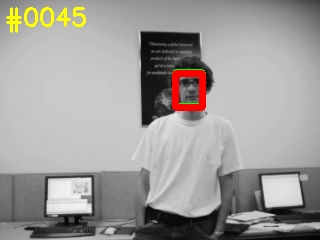}
\hspace{-2pt}
\includegraphics[width=0.16\linewidth,height=0.11\linewidth]{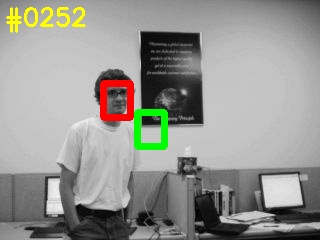}
\hspace{-2pt}
\includegraphics[width=0.16\linewidth,height=0.11\linewidth]{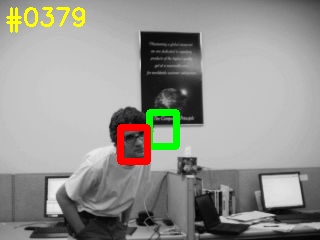}\\
(c) MIL \hspace{210pt} (d) OAB\\
\vspace{4pt}
\includegraphics[width=0.16\linewidth,height=0.11\linewidth]{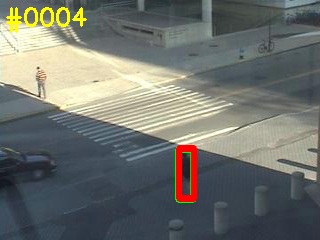}
\hspace{-2pt}
\includegraphics[width=0.16\linewidth,height=0.11\linewidth]{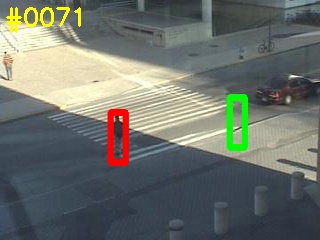}
\hspace{-2pt}
\includegraphics[width=0.16\linewidth,height=0.11\linewidth]{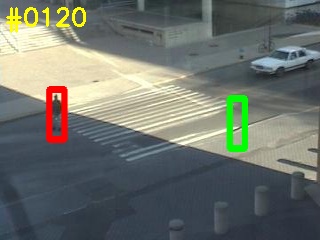}
\hspace{-0pt}
\includegraphics[width=0.16\linewidth,height=0.11\linewidth]{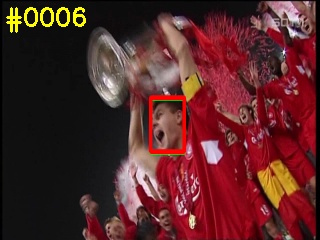}
\hspace{-2pt}
\includegraphics[width=0.16\linewidth,height=0.11\linewidth]{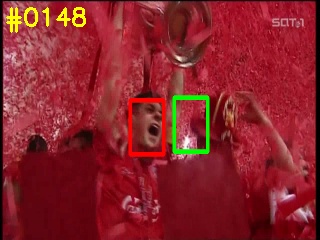}
\hspace{-2pt}
\includegraphics[width=0.16\linewidth,height=0.11\linewidth]{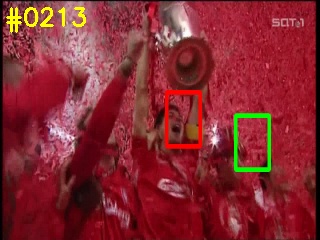}\\
(e) SemiT \hspace{210pt} (f) Struck\\
\vspace{4pt}
\includegraphics[width=0.16\linewidth,height=0.11\linewidth]{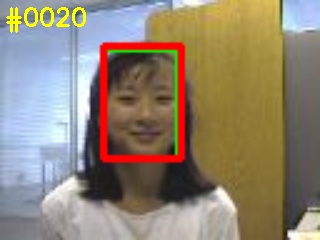}
\hspace{-2pt}
\includegraphics[width=0.16\linewidth,height=0.11\linewidth]{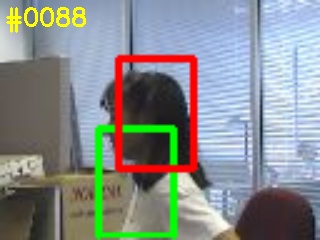}
\hspace{-2pt}
\includegraphics[width=0.16\linewidth,height=0.11\linewidth]{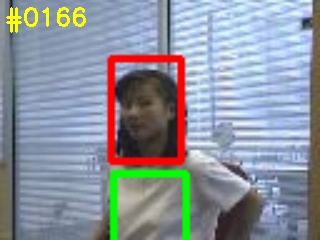}
\hspace{-0pt}
\includegraphics[width=0.16\linewidth,height=0.11\linewidth]{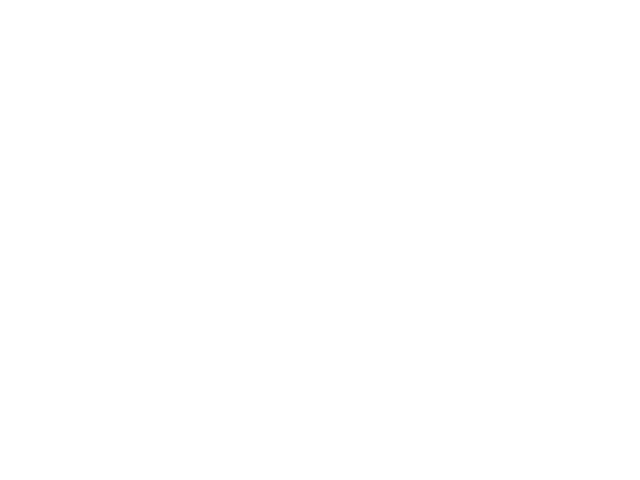}
\hspace{-2pt}
\includegraphics[width=0.16\linewidth,height=0.11\linewidth]{figures/succ-show/white.jpg}
\hspace{-2pt}
\includegraphics[width=0.16\linewidth,height=0.11\linewidth]{figures/succ-show/white.jpg}\\
(g) TGPR \hspace{240pt}
\caption{Examples where adaptive objectness (ADOBING) helps tracking. The results of the base tracker are shown in green, and the results using ADOBING in red. The name of the base tracker is shown under the results. 
}
\label{fig:all-success-show}
\end{figure*}

\begin{figure*}[htbp]
\centering
\subfigure[Failure due to scale variation (tracking a car)]{
\label{fig:failure-lr}
    \includegraphics[width=0.16\linewidth,height=0.119\linewidth]{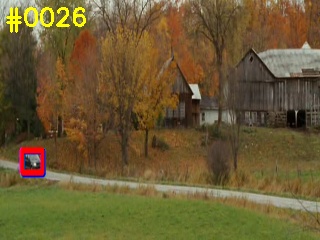}
    \hspace{-3pt}
    \includegraphics[width=0.16\linewidth,height=0.119\linewidth]{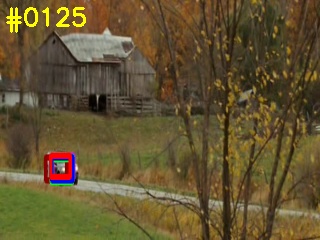}
    \hspace{-3pt}
    \includegraphics[width=0.16\linewidth,height=0.119\linewidth]{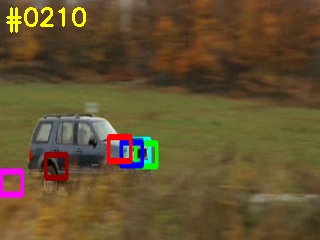}
}
\hspace{-6pt}
\subfigure[Failure due to illumination variation (tracking a body)]{
\label{fig:failure-occ}
    \includegraphics[width=0.16\linewidth,height=0.119\linewidth]{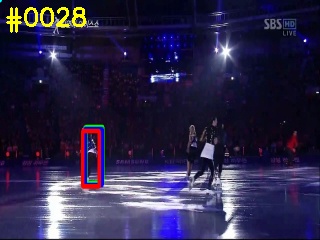}
    \hspace{-3pt}
    \includegraphics[width=0.16\linewidth,height=0.119\linewidth]{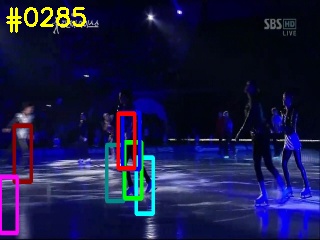}
    \hspace{-3pt}
    \includegraphics[width=0.16\linewidth,height=0.119\linewidth]{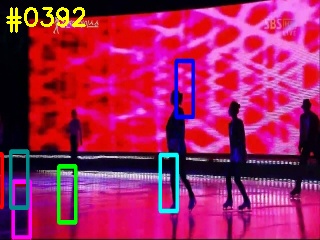}
}\vspace{-1.5mm}\\
\subfigure[Failure due to occlusion(tracking a face)]{
\label{fig:failure-mb}
    \includegraphics[width=0.16\linewidth,height=0.119\linewidth]{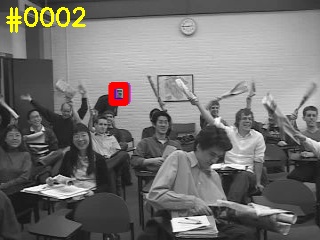}
    \hspace{-3pt}
    \includegraphics[width=0.16\linewidth,height=0.119\linewidth]{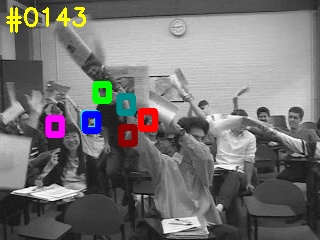}
    \hspace{-3pt}
    \includegraphics[width=0.16\linewidth,height=0.119\linewidth]{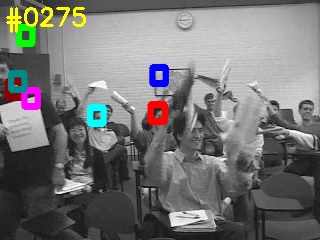}
}
\hspace{-6pt}
\subfigure[Failure due to low resolution (tracking a head)]{
\label{fig:failure-sv}
    \includegraphics[width=0.16\linewidth,height=0.119\linewidth]{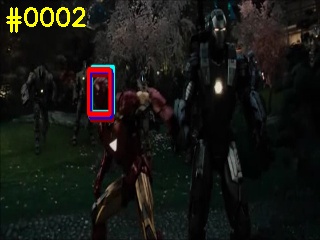}
    \hspace{-3pt}
    \includegraphics[width=0.16\linewidth,height=0.119\linewidth]{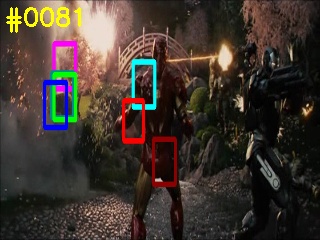}
    \hspace{-3pt}
    \includegraphics[width=0.16\linewidth,height=0.119\linewidth]{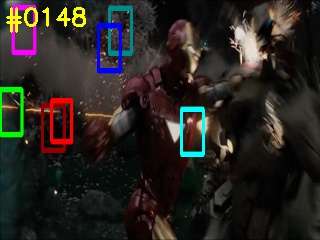}
}\\
\includegraphics[width=0.56\linewidth]{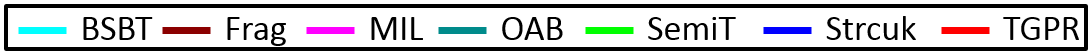}
\caption{Some failures observed in our experiment involving different challenge factors. The legend shows the corresponding base trackers.}
\label{fig:failure}
\end{figure*}

\vspace{-2.501mm}\paragraph{Results.}
Figure~\ref{fig:trackers-compare-cvpr2013} shows the success and the precision plots of the seven ADOBING-enhanced trackers and their corresponding base trackers. In both plots, we can see that the ADOBING-enhanced trackers are consistently better than  their corresponding base ones. Table~\ref{table:auc-compare-cvpr2013} gives the quantitative comparison of the base trackers and the two versions of objectness-enhanced trackers.  The results show that the proposed adaptive objectness (i.e., ADOBING) brings more benefits than BING for all base trackers. Figure~\ref{fig:all-success-show} shows example tracking results where ADOBING helps tracking.

It is worth noting that the ADOBING-enhanced TGPR (precision score for ranking: 0.7678) outperforms previously reported best ones - 0.732 by~\cite{henriques2014high} and 0.733 by~\cite{gao2014transfer}.

It is unrealistic to expect that tracking can be entirely solved by  integrating objectness. That said, it is useful to investigate the failures to better understand the proposed trackers. Figure~\ref{fig:failure} shows some typical failures observed in our experiments. Roughly speaking, these failures are mainly due to challenges that bother most existing trackers. 

\vspace{-2.501mm}\paragraph{Attribute-based performance analysis.}
Taking advantage of the attribute annotation of each sequence, we analyze the performance of the object-adaptive objectness for visual tracking under different challenging factors. Table~\ref{table:att-gain-cvpr2013} summarizes the performance gain of using ADOBING. The AUC for base trackers under each attribute is calculated by averaging the AUC of all the base trackers on the corresponding subset of sequences; the AUC for the ADOBING-enhanced trackers is obtained in a similar way.

From the results, we can see that ADOBING helps visual tracking algorithms consistently under all the challenge factors. The largest performance gain is achieved for out-of-view (OV). A possible reason for this is that when an object is moving out of view, it often generates partial objects to incorrectly update the base tracker; by contrast, ADOBING helps inhibiting such partial objects since they usually have low objectness. On the other end, the gain for low resolution (LR) is relatively small, which can be attributed to the lack of reliable guidance from the base trackers due to the weak appearance information.

\begin{table*}[!t]
\caption{Evaluation results on the Princeton Tracking Benchmark. }
\centering
\small
\begin{tabular}
{c|c|c|c|c|c|c|c|c|c|c|c|c}
\hline
    \multirow{2}{*}{Algorithm}&\multirow{2}{*}{\tabincell{c}{overall\\SR}} & \multicolumn{3}{c|}{target type} &\multicolumn{2}{c|}{target size} & \multicolumn{2}{c|}{movement} &\multicolumn{2}{c|}{occlusion}&\multicolumn{2}{c}{motion type}\\
\cline{3-13}
&&human&animal&rigid&large&small&slow&fast&yes&no&passive&active\\
\hline
TGPR+ADOB&0.489&0.39&0.51&0.59&0.47&0.50&0.63&0.43&0.37&0.65&0.55&0.46\\
\rowcolor{gray!25}TGPR&0.472&0.36&0.51&0.58&0.46&0.48&0.62&0.41&0.35&0.65&0.56&0.44\\
Struck+ADOB&0.447&0.38&0.47&0.52&0.44&0.45&0.58&0.40&0.30&0.65&0.52&0.42\\
\rowcolor{gray!25}Struck&0.444&0.35&0.47&0.53&0.45&0.44&0.58&0.39&0.30&0.64&0.54&0.41\\
Frag+ADOB&0.429&0.38&0.43&0.49&0.45&0.41&0.62&0.35&0.35&0.54&0.49&0.41\\
\rowcolor{gray!25}Frag&0.412&0.39&0.41&0.44&0.46&0.37&0.58&0.35&0.33&0.52&0.46&0.39\\
OAB+ADOB&0.405&0.32&0.44&0.49&0.38&0.43&0.53&0.35&0.29&0.57&0.51&0.37\\
\rowcolor{gray!25}MIL+ADOB&0.403&0.32&0.51&0.44&0.39&0.41&0.54&0.35&0.29&0.56&0.48&0.37\\
SemiT+ADOB&0.385&0.34&0.42&0.42&0.37&0.40&0.52&0.33&0.30&0.51&0.49&0.34\\
\rowcolor{gray!25}OAB&0.382&0.28&0.46&0.46&0.35&0.41&0.52&0.32&0.27&0.54&0.48&0.35\\
MIL&0.355&0.32&0.37&0.38&0.37&0.35&0.46&0.31&0.26&0.49&0.40&0.34\\
\rowcolor{gray!25}BSBT+ADOB&0.296&0.25&0.26&0.37&0.31&0.29&0.43&0.24&0.26&0.35&0.43&0.25\\
BSBT&0.285&0.22&0.30&0.36&0.27&0.29&0.42&0.23&0.26&0.32&0.43&0.23\\
\rowcolor{gray!25}SemiT&0.283&0.22&0.33&0.33&0.24&0.32&0.38&0.24&0.25&0.33&0.42&0.23\\
\hline
\end{tabular}
\label{table:ptb-ranking}
\end{table*}

\begin{table*}[!t]
\small
\caption{The performance of BING- or ADOBING- enhanced trackers and their corresponding base trackers in term of \emph{success rate} on PTB. The \textcolor{red}{\bf best} and \textcolor{blue}{second best} for each tracker are indicated by \textcolor{red}{\bf red} and \textcolor{blue}{blue} respectively.}
\centering
\begin{tabular}{|c|c|c|c|c|c|c|c||c|}
\hline
&BSBT \cite{stalder2009beyond}&Frag \cite{adam2006robust}&MIL \cite{babenko2011robust}&OAB \cite{grabner2006real}&SemiT \cite{grabner2008semi}&Struck \cite{hare2011struck}&TGPR \cite{gao2014transfer}&Average\\
\hline
Base&0.285&0.412&0.355&0.382&0.283&\textcolor{blue}{0.444}&0.472&0.376\\
\hline
Base+BING&\textcolor{red}{\bf 0.305}&\textcolor{red}{\bf 0.434}&\textcolor{red}{\bf 0.408}&\textcolor{blue}{0.388}&\textcolor{blue}{0.367}&0.434&\textcolor{red}{\bf 0.505}&\textcolor{blue}{0.406}\\
\hline
Base+ADOB&\textcolor{blue}{0.296}&\textcolor{blue}{0.429}&\textcolor{blue}{0.403}&\textcolor{red}{\bf 0.405}&\textcolor{red}{\bf 0.385}&\textcolor{red}{\bf 0.447}&\textcolor{blue}{0.489}&\textcolor{red}{\bf 0.408}\\
\hline
\end{tabular}
\label{table:sc-compare-ptb}
\end{table*}

\subsection{Princeton Tracking Benchmark}
The recently proposed Princeton Tracking Benchmark (PTB) \cite{song2013tracking} contains 100 RGBD sequences divided into 11 categories. Among these 100 sequences, the ground truth of 5 of them are released for parameter tuning and the rest are withheld for evaluation. Though the purpose of this dataset is to evaluate RGBD trackers, the available RGB sequences and the evaluation  website\footnote{http://vision.princeton.edu/projects/2013/tracking/} make it suitable to verify the usefulness of the objectness for visual tracking. Note that due to limitation of current depth acquisition techniques, all the sequences are captured indoors.
We follow the protocol in~\cite{song2013tracking} and submit the results to the PTB website for evaluation.  The success rate is calculated by thresholding the overlap between the tracked bounding box of the target and the ground truth.

Table~\ref{table:ptb-ranking} summarizes  the overall success rate and the success rate under each category of the evaluated tracker. For the base trackers MIL, SemiT, Struck which have already been evaluated in~\cite{song2013tracking}, we use the most recent results from the evaluation website directly. From the results, we can see that all seven base trackers can benefit from integrating the proposed adaptive objectness, an observation consistent with our experiments on the CVPR2013 benchmark.

Table~\ref{table:sc-compare-ptb} lists the comparison of the base tracker with the two versions of objectness-enhanced versions. On one hand, it again confirms the consistent improvement using the proposed ADOBING objectness; on the other hand, it shows that the improvement using ADOBING is similar to that using the original BING. That said, BING is less stable since it actually hurts the performance when using Struck as the base tracker.

\section{Conclusion}
\label{sec:conclusion}

In this paper, we propose to use adaptive objectness for assisting object tracking. Based on the recently proposed fast objectness algorithm named BING, we have designed a tracking-adaptive objectness named ADOBING through adaptive SVM. ADOBING effectively adjust the general objectness estimation for taking into consideration tracking specific information. Consequently, when integrated into a base tracker, it can help reduce the chance of drifting by avoiding tracking candidates that do not appear like an object. To validate the idea, ADOBING is integrated into seven highly ranked trackers chosen from recent published evaluations. Then these trackers are tested on two public benchmarks including in total 150 sequences. The results show that integration of ADOBING, even in a straightforward way, consistently improves these trackers.

{\small
\bibliographystyle{ieee}
\bibliography{egbib}
}

\end{document}